\documentclass[a4paper, 10pt, conference]{ieeeconf}
\usepackage{graphicx}%
\usepackage{multirow}%
\usepackage{amsmath,amssymb,amsfonts}%
\usepackage{mathrsfs}%
\usepackage{xcolor}%
\usepackage{textcomp}%
\usepackage{manyfoot}%
\usepackage{booktabs}%
\usepackage{algorithm}%
\usepackage{algorithmicx}%
\usepackage{algpseudocode}%
\usepackage{listings}%
\usepackage{placeins}
\IEEEoverridecommandlockouts

\title{A Model of the Sidewalk Salsa}

\author{Olger Siebinga$^{1}$%
\thanks{$^{1}$\textit{Department of Cognitive Robotics, Delft University of Technology}, Delft, The Netherlands {\tt\small o.siebinga@tudelft.nl}}%
\thanks{\textbf{Data and Software Availability}
The software to simulate the model was implemented in Python with Casadi~\cite{Andersson2018} and is published online: {\tt\small https://github.com/tud-hri/sidewalk-simulation}. The data underlying this publication is published on the 4TU data repository~\cite{Siebinga2024b}.}}

\begin{document}

\maketitle
\thispagestyle{empty}
\pagestyle{empty}

\begin{abstract}
When two pedestrians approach each other on the sidewalk head-on, they sometimes engage in an awkward interaction, both deviating to the same side (repeatedly) to avoid a collision. This phenomenon is known as the sidewalk salsa. Although well known, no existing model describes how this “dance” arises. Such a model must capture the nuances of individual interactions between pedestrians that lead to the sidewalk salsa. Therefore, it could be helpful in the development of mobile robots that frequently participate in such individual interactions, for example, by informing robots in their decision-making. Here, I present a model based on the communication-enabled interaction framework capable of reproducing the sidewalk salsa. The model assumes pedestrians have a deterministic plan for their future movements and a probabilistic belief about the movements of another pedestrian. Combined, the plan and belief result in a perceived risk that pedestrians try to keep below a personal threshold. In simulations of this model, the sidewalk salsa occurs in a symmetrical scenario. At the same time, it shows behavior comparable to observed real-world pedestrian behavior in scenarios with initial position offsets or risk threshold differences. Two other scenarios provide support for a hypothesis from previous literature stating that cultural norms, in the form of a biased belief about on which side others will pass (i.e. deviating to the left or right), contribute to the occurrence of the sidewalk salsa. Thereby, the proposed model provides insight into how the sidewalk salsa arises.
\end{abstract}

\section{Introduction}
\begin{figure*}[ht]
    \centering
    \includegraphics[width=1.\textwidth]{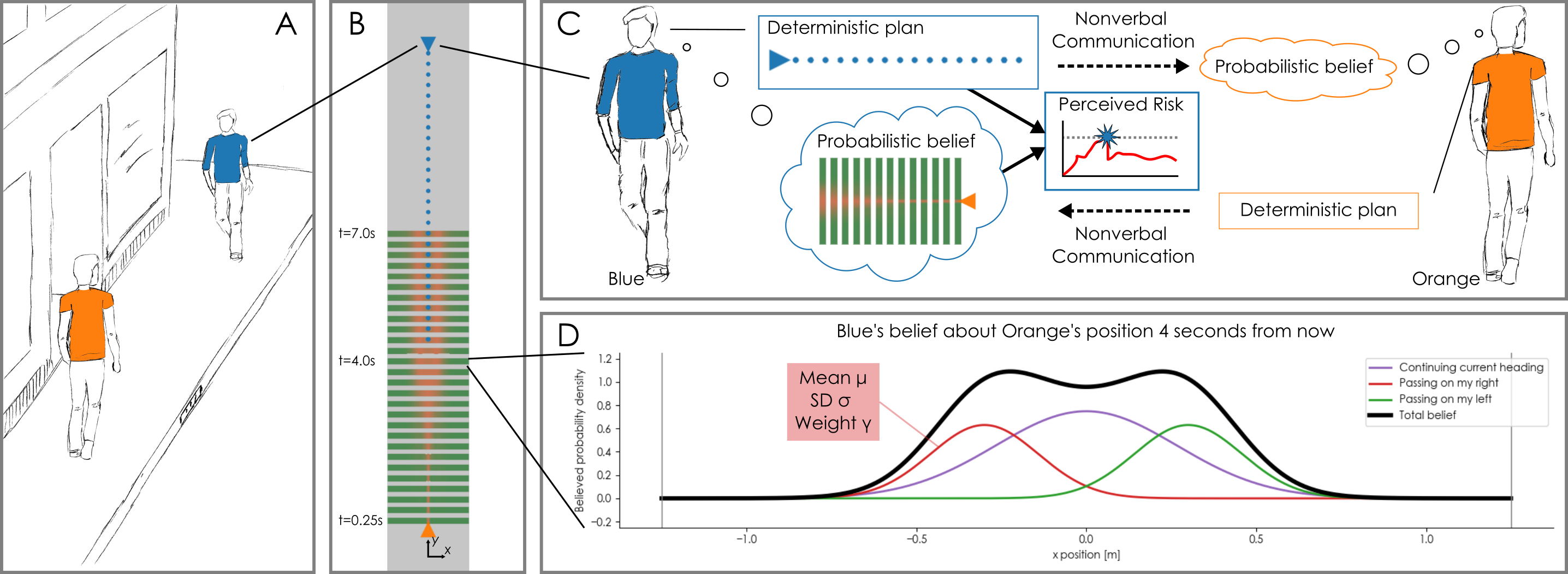}
    \caption{Panel \textbf{A}: a sketch of the situation of interest: two pedestrians approaching each other on a sidewalk. Panel \textbf{B}: a top-down view of the modeled environment. The orange and blue triangles represent the pedestrians' positions and headings. Panel \textbf{C}: a schematic overview of the model. Every pedestrian is assumed to have a deterministic plan for their future trajectory --represented by dots-- and a probabilistic belief about the future trajectory of the other pedestrian, represented by the heat map where orange represents a high and green a low believed probability density. Combined, the plan and belief lead to a perceived risk; if this risk exceeds a personal threshold, pedestrians update their plan to lower it. The belief is based on nonverbal communication (i.e., the movements) from the other pedestrian. Panel \textbf{D}: The belief takes the form of a set of probability distributions over the lateral (x) position of the other pedestrian at a specific longitudinal (y) position. Each slice represents a probability distribution, combining three Gaussian distributions, each representing a specific high-level behavior of the other pedestrian: continuing on their current heading, passing on the ego's right side, and passing on the ego's left side. Each Gaussian distribution is defined by a mean $\mu$ and standard deviation $\sigma$ and is scaled by a weight $\gamma$. The three weights always sum to one.}
    \label{fig:introduction-overview}
\end{figure*}

While walking down the sidewalk, you're approaching another pedestrian head-on (Figure~\ref{fig:introduction-overview}-A). To avoid bumping into them, you move to your right. While you're doing that, the other pedestrian moves to their left, moving into your path again. In response, you slow down and move to your left, but again, the other pedestrian moves to their right simultaneously. Eventually, you use a hand gesture to indicate where they can pass, and you are both on your way again. Although most people won't encounter such an awkward interaction on a daily basis, it is undoubtedly familiar, as evidenced by the numerous names under which this phenomenon is known: the sidewalk dance~\cite{Holohan2018, Bugler1997}, the footpath foxtrot~\cite{Burgress2019}, the pavement polka~\cite{LaBelle2015b}, the I’m-trying-to-get-around-you dance~\cite{Munro2019}, the sidewalk shuffle~\cite{Holohan2018, Burgress2019}, or --under the name I'm using here-- the sidewalk salsa~\cite{Burgress2019, Munro2019, LaBelle2015a}.

The sidewalk salsa represents a negotiation between two pedestrians to find a safe solution to a spatial conflict. It involves the pedestrians' expectations about what the other person will do~\cite{Burgress2019} influenced by social norms~\cite{Holohan2018, Munro2019} and individual preferences in terms of personal space~\cite{Munro2019}. Pedestrians' motions can be interpreted as communication about their future plans. And somehow, in the case of the sidewalk salsa, these interpretations and beliefs about others lead to a (recurring) conflict. The awkward sidewalk salsa thus encapsulates many nuances of interactions between individual pedestrians in a single phenomenon. 

A good understanding of the sidewalk salsa would increase our fundamental knowledge of the nuances of pedestrian interactions in general, such as how pedestrians form expectations (a belief) about other people's future paths. Understanding these nuances is necessary to enable the design and development of efficient and safe automated behavior for robots operating in environments where they interact with individual pedestrians; for example, mobile parcel delivery robots (e.g.,~\cite{Gehrke2023, Weinberg2023}) or indoor mobile robots (e.g.,~\cite{Dadvar2021, Chen2018}) which are used in healthcare or industrial environments~\cite{Fang_Mei_Yuan_Wang_Wang_Wang_2021, Möller_Furnari_Battiato_Härmä_Farinella_2021}. The development of such robotic behaviors would benefit from a valid model of the sidewalk salsa. Such a model could not only increase our understanding~\cite{Guest2021} of the internal beliefs and expectations of pedestrians and how they are formed but could also inform robots online about the intentions of humans (as is done in autonomous vehicles~\cite{Siebinga2022}).

Although scientists have said much about the sidewalk salsa in newspapers and on websites (e.g.,~\cite{Bugler1997, Burgress2019, Holohan2018, Munro2019}), it is not yet a common topic in the scientific literature. In 2015, Honma et al. studied a similar phenomenon (which they called "hesitant avoidance while walking") in a lab experiment~\cite{Honma2015}. They observed this behavior but did not model it. The main difference with the sidewalk salsa, where two pedestrians step into each other's path, was that they also regarded hesitant behavior in a single pedestrian. Nummenmaa et al. also investigated how conflicts in head-on encounters appear by recording eye gaze data in an experiment where a participant encountered an animated human~\cite{Nummenmaa2009}. They found that the gaze direction of an animated pedestrian made the participant look and go in the other direction, showing that non-verbal cues play an important role in the beliefs about another pedestrian's plan.

Other studies on individual pedestrian behavior mostly focus on pedestrian preferences in terms of personal space. They aim to identify the distance pedestrians keep to moving or static obstacles and humans in a laboratory~\cite{Gorrini2014, Olivier2012, Gerin-Lajoie2005, Vassallo2018, Pfaff2018, Moussaid2009}. This personal-space-based approach does not consider interactions (e.g., communication) and expectations between pedestrians. Furthermore, these approaches mostly use a social forces model (e.g.,~\cite{Moussaid2009}) or a constant velocity model (e.g.,~\cite{Pfaff2018}) to determine other pedestrians' actions. Therefore, pedestrians are assumed to respond to the current state of the world and do not take multiple high-level actions (i.e., passing on my left \underline{or} right side) of others into account. In some experiments, this restriction of high-level choices is ensured in the experiment design by restricting the possibilities of (one of) the pedestrians~\cite{Vassallo2017, Pfaff2018}.

These same underlying mechanisms of force field and constant velocity are also used in many studies of pedestrian and robot interactions in larger groups of pedestrians (i.e., crowds) in the real world~\cite{Lerner2007, Pellegrini2009, Biswas2022, Hoogendoorn_Daamen_2005} or models~\cite{Echeverria-Huarte2023, Lerner2007, Pellegrini2009, Brito2020}. These do not focus on interactions between individual participants or describe the nuances of these individual interactions. A model of one-on-one sidewalk interactions that describes the sidewalk salsa and thus captures the underlying beliefs and expectations of pedestrians is missing from the literature. 

In this work, I present a model of one-on-one pedestrian interactions on a sidewalk (Figure~\ref{fig:introduction-overview}-B \& C) capable of reproducing the sidewalk salsa under specific circumstances. The model is based on the Communication-Enabled-Interaction (CEI) framework~\cite{Siebinga2023}, a framework developed to model traffic interactions. With the model, I aim to explore the hypothesis that the occurrence of the sidewalk salsa is partly due to a cultural norm prescribing on which side pedestrians pass each other~\cite{Munro2019, Holohan2018, Pfaff2018, Moussaid2009}. This is based on the idea that in countries where cars drive on the right side of the road, pedestrians would be more likely to move to the right side of the sidewalk. In the model, this cultural norm is implemented as a bias in the belief about on which side the other pedestrian will pass. The results show that when pedestrians have matching belief biases, the number of sidewalk salsas decreases while opposing belief biases cause an increase in sidewalk salsas.

\section{Methods}
The model was simulated in a custom Python environment with five scenarios to investigate the circumstances under which the sidewalk salsa appears.

\subsection{Simulation Environment}
The simulated environment consisted of a sidewalk of $2.5~m$ wide and $15.0~m$ long (Figure~\ref{fig:introduction-overview}-B). Pedestrians move in the 2-dimensional x-y-plane and adhere to a pedestrian dynamics model developed by Mombaur et al. in 2009~\cite{Mombaur_Truong_Laumond_2010}. This dynamics model includes non-holonomic motions, which were previously found to be representable of human trunk motion during walking~\cite{Arechavaleta2008}. However, it also allows for holonomic motions in the form of side-stepping, which could be an important factor in the sidewalk salsa. The model expresses the accelerations and velocities in a local coordinate frame attached to the body, with a forward ($a_{forw}, v_{forw}$), orthogonal ($a_{orth}, v_{orth}$), and angular ($\dot{\omega}, \omega$) component. The pedestrians are assumed to directly control the accelerations, with maximum absolute values of $a_{forw}=2.0~m/s^2, a_{orth}=1.0~m/s^2, \dot{\omega}=\pi~rad/s^2$. The simulation ran at $20~Hz$. The pedestrians started on opposite sides of the sidewalk, with a heading aligned with the sidewalk and an initial velocity of $1.3~m/s$ (an ordinary/medium walking speed~\cite{Zebala2012, Gorrini2014}). The simulation finished when one of the pedestrians reached the opposite end of the track. It was aborted prematurely if the distance between the pedestrians' (center) positions was less than $0.25~m$ (a collision) or when one of the pedestrians moved beyond the side of the track (out-of-bounds).

\subsection{Model}
The model of pedestrian interactions is based on the Communication-Enabled-Interaction (CEI) framework~\cite{Siebinga2023}, developed to create models of traffic interactions and previously for vehicle interactions~\cite{Siebinga2023, Siebinga2024}. The CEI framework assumes all agents (pedestrians) have a deterministic plan for their future trajectory. They combine this plan with a probabilistic belief about the future movements of other agents based on communication they receive (Figure~\ref{fig:introduction-overview}-C). This combination of plan and belief leads to a perceived risk, which is constantly evaluated. If the risk exceeds a personal threshold, agents are assumed to update their plan to lower their perceived risk; otherwise, they continue their current plan. The model consists of implementations for each component from the CEI framework.

\paragraph{Plan}
The plan of a pedestrian consists of a set of waypoints over a time horizon $T=7.0~s$ (Figure~\ref{fig:introduction-overview}-B) with a frequency of $4~Hz$. The plan is obtained by minimizing a cost function $c$:
\begin{equation}
\begin{aligned}
    c &= \lambda_1 dv_{forw}^2 + \lambda_2 (v_{forw}^{-})^2 + \lambda_3 v_{orth}^2 + \lambda_4 d\theta^2 + \\
    & \lambda_5 \dot{\omega}^2 + \lambda_6 a_{forw}^2 + \lambda_7 a_{orth}^2,
\end{aligned}
\end{equation}
where $dv_{forw}$ is the deviation from the initial forward velocity, $v_{forw}^{-}$ is the negative forward velocity (i.e., it is only non-zero when the forward velocity is negative to penalize walking backward), $v_{orth}$ is the orthogonal velocity (i.e., side-stepping), $d\theta$ is the deviation from the initial heading, $a_{forw}, a_{orth}$ and $\dot{\omega}$ are the normalized accelerations. The $\lambda$-parameters are scaling factors ($\lambda_1=\lambda_5=\lambda_6=\lambda_7=1$, $\lambda_2=100$, $\lambda_3=2$, $\lambda_4=5$). 

This cost function only contains terms regarding getting to the other end of the sidewalk quickly (deviations from velocity and heading) and comfortably (inputs). Safety is imposed by constraining the perceived risk. The optimization is performed once at the start of the simulation to obtain an initial plan (without a risk constraint). It is then repeated when the perceived risk of the pedestrian exceeds their personal threshold $\rho$. In that case, the perceived risk for the new plan is constrained to  ($0.75\rho$). If the optimization algorithm fails, the pedestrian is assumed to slow down and come to a full stop in half a second ($a=-2v$ for all accelerations). The optimization is repeated at the next timestep with a looser risk constraint ($0.9\rho$) to find a feasible plan again quickly. 

At every simulated timestep, the first input from the (updated) plan is applied and removed from the plan. The last step is duplicated, representing "continuing the current plan."~\cite{Siebinga2023}

\paragraph{Communication}
Pedestrians base their belief about the other's future trajectory on communication through observed movements. At every timestep, pedestrians are assumed to observe the other pedestrian's position without error. Velocity and heading are subject to noise and a perception delay based on an update rate (based on~\cite{Siebinga2024}). The observed (subscript $o$) velocity and heading are updated based on the error with the current true (subscript $t$) values. The updates are defined as:
\begin{align}
    \Delta v_{forw, o} &= \alpha (v_{forw, t} - v_{forw, o}) + \beta \epsilon_1 \\
    \Delta v_{orth, o} &= \alpha (v_{orth, t} - v_{orth, o}) + \beta \epsilon_2 \\
    \Delta \theta_o &= \alpha (\theta_t - \theta_o) + \beta \epsilon_3 \\
    \epsilon_n &\sim \mathcal{N}(\mu=0.0, \sigma=\sqrt{dt}=\sqrt{0.05}).
\end{align}
In these equations, $\alpha$ denotes the update rate of the perception ($\alpha=2 * dt$), representing that pedestrians need $0.5$ seconds to detect changes. $\epsilon_n$ is noise drawn from a normal distribution, scaled by $\beta=0.03$, resulting in a walking (Brownian) noise in the believed velocities and heading. The observed forward velocity is constrained to be positive to prevent sign issues in the belief construction.

\paragraph{Belief}
The belief consists of belief points subject to the same time horizon ($7.0~s$) and planning frequency ($4~Hz$) as the plan (Figure~\ref{fig:introduction-overview}-B). Each belief point is placed at a fixed longitudinal (y) position, assuming a constant observed y-velocity (in world-frame) of the other pedestrian. The belief point represents a believed probability distribution over the lateral (x) position for the other pedestrian, consisting of three high-level belief components, each described by a Gaussian distribution ($\mathcal{N}(\mu, \sigma)$) with a weight ($\gamma$) (Figure~\ref{fig:introduction-overview}-D).

The first belief component represents the possibility that the other pedestrian will continue with their current heading (denoted by subscript $c$). The weight of this belief component is always the same: $\gamma_{c} = 0.5$. The mean is calculated by extrapolating the current observed position, velocity, and heading of the other pedestrian. The standard deviation is based on an expected lateral acceleration $a_e = 0.2~m/s^2$, and therefore increases with time until the belief point ($t_b$):
\begin{equation}
    \sigma_{c,t_b} = \frac{1}{2} \frac{a_e}{3} t_b^2.
\end{equation}
The factor $\frac{1}{3}$ comes from the probability of 0.997 that values fall within the range $+-3\sigma$. 
The other two belief components represent the cases where the other pedestrian will pass on the ego's left (subscript $l$) or right (subscript $r$) side. Their weights are proportional to the current observed position and velocity of the other pedestrian with respect to the ego pedestrian's (subscript $e$) position:
\begin{align}
    \gamma_{l} &= (1 - \gamma_{c}) * \nonumber \\ 
               &(\frac{1}{2} + \frac{1}{\pi}(atan\left( \frac{\zeta(y_o - y_e)}{x_o - x_e} \right) - \theta_{e,i}) - v_{orth, o}), \\
    \gamma_{r} &= 1 - \gamma_{c} - \gamma_{l},
\end{align}
where $theta_{e,i}$ denotes the ego pedestrian's initial heading, and $\zeta=\frac{1}{4}$ is a scaling factor. Both $\gamma_{l}$, and $\gamma_{r}$ are constrained between $(1.0 - \gamma_{c}, 0.0)$. These equations describe that if the other pedestrian moves to the right side of the sidewalk (from the ego pedestrian's perspective) and gets closer to them, the probability (weight) that they will pass on the right side increases. A larger distance between the pedestrians increases the believed probability that the other pedestrians will change sides. At the same time, an orthogonal velocity (i.e., side stepping) to one side also increases the believed probability that the other pedestrian will pass on that side. 
The minimal comfortable range ($r_{com}=0.3~m$) represents the range within which pedestrians do not tolerate others. The value is taken from an observed distribution of lateral spacing in real-world data~\cite{Kim2014} ($r_{com}=\mu-1\sigma$). This parameter helps to determine the means and standard deviations:
\begin{equation}
\mu_{l/r, t_b} = \begin{cases}
                x_e +- r_{com}, & \text{if } abs(x_{o, t_b} - x_e) < r_{com}\\
                x_{o, t_b},              & \text{otherwise}
                \end{cases},
\end{equation}
where $x_e$ denotes the current x position of the ego pedestrian, and $l/r$ is determined by $+-$ depending on the perspective of the ego pedestrian. This equation reflects that the mean of the belief component representing a pass on the left or right is equal to the mean of the component representing the continuation of the current heading if the current heading of the other pedestrian steers clear of the comfortable range. In other cases, the belief component represents that the other pedestrian will pass at a minimum distance. The standard deviations are determined based on the currently available space on the left ($dx_l$) and right ($dx_r$) side of the ego pedestrian. These are defined as:
\begin{align}
    dx_l &= abs(x_{lb} - x_e),~dx_r = abs(x_{rb} - x_e) \\
    \sigma_{l, t_b} &= \frac{1}{6}(dx_l - r_{com}),~
    \sigma_{r, t_b} = \frac{1}{6}(dx_r - r_{com}),
\end{align}
where $x_{lb}$ and $x_{rb}$ denote the x-location of the left and right bound of the sidewalk from the ego pedestrian's perspective. Combined, these three belief components form the belief probability distribution for the lateral position of the other pedestrian at a specific point in time ($t_b$) and a fixed longitudinal position (Figure~\ref{fig:introduction-overview}-D). The weights $\gamma$ of the three components always sum to one. 

\paragraph{Risk Perception}
Two elements contribute to risk perception: getting too close to the other pedestrian ($\mathrm{P}_{close}$) and going beyond the bounds of the sidewalk ($\mathrm{P}_{bounds}$) each ranging from $[0, 1]$. Every corresponding plan-belief point pair is evaluated on both aspects and these two contributions are summed. The maximum risk over all points is the pedestrian's perceived risk ($\mathrm{P}$). This perceived risk can technically range from $[0-2]$. However, the personal risk threshold $\rho$ should always be on the interval $[0-1]$ to ensure that risk from a single source can trigger a replan. The rationale behind this is that the replanning mechanism is ignorant of the source of the risk.

The risk of getting too close to the other pedestrian is based on the combination of the believed probability of the lateral positions of the other pedestrian and the ego pedestrian's deterministic planned position at the same point in time. The probability that the other pedestrian will be within the lateral comfortable range of the ego pedestrian at the belief time point $t_b$ can be calculated:
\begin{equation}
p(x_{e,t_b} - r_{com}<x_{b, t_b}<x_{e,t_b} + r_{com}),     
\end{equation}
where $x_{b, t_b}$ is the belief x-position of the other pedestrian at time $t_b$. This probability is then scaled by a factor $f_y$ to account for the longitudinal distance between the belief and plan point ($dy$, which is deterministic). This factor is defined as:
\begin{align}
    f_y &= exp(-dy^2 / (2 r_{com}) ^ 2))
\end{align}
The risk of going beyond the bounds of the sidewalk is evaluated based on the planned positions alone. Two steep Sigmoid functions, one for each side of the track, represent the risk of leaving the sidewalk. A time discount factor $f_t$ lowers the perceived risk for plan points further away:
\begin{align}
    f_t &= e^{-t_b/T} \\
    \mathrm{P}_{bounds} &= f_t (1 - \frac{1}{2}tanh(\eta (x_{e,t_b} + (x_{lb} - \delta x)) + \\
    &\frac{1}{2}tanh(\eta(x_{e,t_b} - (x_{rb} + \delta x))),
\end{align}
in these equations, $\eta=10$ is a scaling factor that determines the "steepness" of the Sigmoid functions, and $\delta x = 0.15~m$ represents a small offset to let the risk be approximately $1$ at the sidewalk bound (instead of the middle of the Sigmoid being aligned with the bound).

\subsection{Scenarios}
Five scenarios were used to test if --and under which circumstances-- the model can reproduce the sidewalk salsa (Table~\ref{tab:scenarios}). In the \textbf{symmetric} scenario, two pedestrians with equal risk thresholds approach each other on the sidewalk's center. It can be expected that the sidewalk salsa will occur here —- however, such a perfectly symmetrical encounter is not very realistic. Two aspects of real-world interactions that could potentially lower the number of sidewalk salsas are investigated here. First, an initial lateral positional difference is tested in the scenario \textbf{different sides}. Second, personal differences between pedestrians are tested in the \textbf{different risk thresholds} scenario. 

\begin{table}[ht!]
    \centering
    \caption{The five simulated scenarios and their parameters.}
    \label{tab:scenarios}
    \begin{tabular}{|p{2.8cm}|p{1.4cm}|p{1.5cm}|p{1.2cm}|}\hline
         Scenario & Risk thresholds $\rho$ & Initial x position offset & Belief bias\\ \hline
         Symmetric & 0.65 / 0.65 &- & - \\ 
         Different sides & 0.65 / 0.65 & $0.1/-0.1~m$ & \\ 
         Different risk thresholds & 0.6 / 0.7 & - & - \\ 
         Same belief bias & 0.65 / 0.65 & - & right/right\\ 
         Different belief bias & 0.65 / 0.65 & - & left/right\\ \hline
    \end{tabular}
\end{table}

Beyond these spatial and personal differences, I explore the hypothesis that cultural norms could play a role in sidewalk salsa~\cite{Holohan2018, Munro2019, Pfaff2018, Moussaid2009}. More specifically, in countries where vehicles drive on the right side of the road, pedestrians could believe that others are more likely to pass on their left -- and vice versa. To reflect this in the model, the weights for the belief components representing the beliefs for passing on the right or left ($\gamma_l, \gamma_r$) can be biased. In the \textbf{same belief bias} scenario, both pedestrians believe that the other pedestrian is more likely to pass on their left ($1.3\gamma_l$) than on their right ($0.7\gamma_r$). The \textbf{different belief bias} scenario uses the same scaling factors but for opposing sides. Note that in these scenarios, the belief weights are the only things that changed in the model; the planning part remained the same. Pedestrians are assumed to have other expectations (based on cultural norms) but do not plan their behavior differently.

\section{Results}
\begin{figure}[h]
    \centering
    \includegraphics[width=\linewidth]{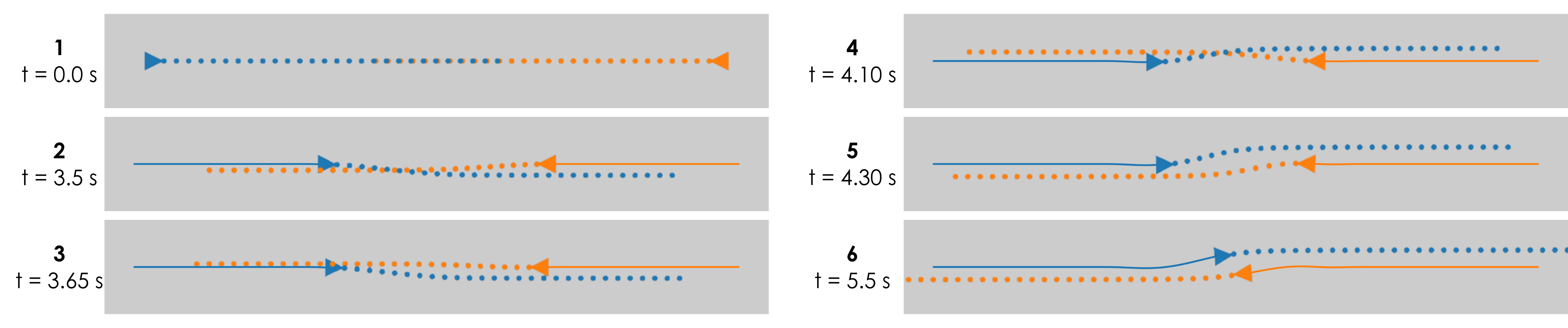}
    \caption{A typical interaction in the symmetric scenarios. Triangles indicate the positions and headings of the orange and blue pedestrians. The dots indicate their planned positions until the time horizon ($T=7.0~s$). The lines show their trajectories up to this point. The text above the images indicates the time in the simulation.}
    \label{fig:example_interaction}
\end{figure}

The model was simulated 100 times in each scenario, resulting in 500 trials. Figure~\ref{fig:example_interaction} shows a typical simulation of the symmetric scenario (frame 1). In this case, the pedestrians ended up in the sidewalk salsa. Initially, only the blue pedestrian decided to deviate from their path (frame 2). The orange pedestrian noticed and changed their plan to move slightly to the right (frame 3). However, the blue pedestrian was under the impression that Orange would move to their right and also changed their plan, at this point both pedestrians decided to move to the same side of the sidewalk (frame 4). So far, the pedestrians have not deviated much from the sidewalk's center and only use minor changes in their heading angle. After some time, the orange pedestrian notices they are heading the same way and changes their plan again to avoid a collision (frame 5). Eventually, Orange passes Blue on their right side, and both pedestrians are on their way again (frame 6). 

\begin{figure*}
    \centering
    \includegraphics[width=1.0\textwidth]{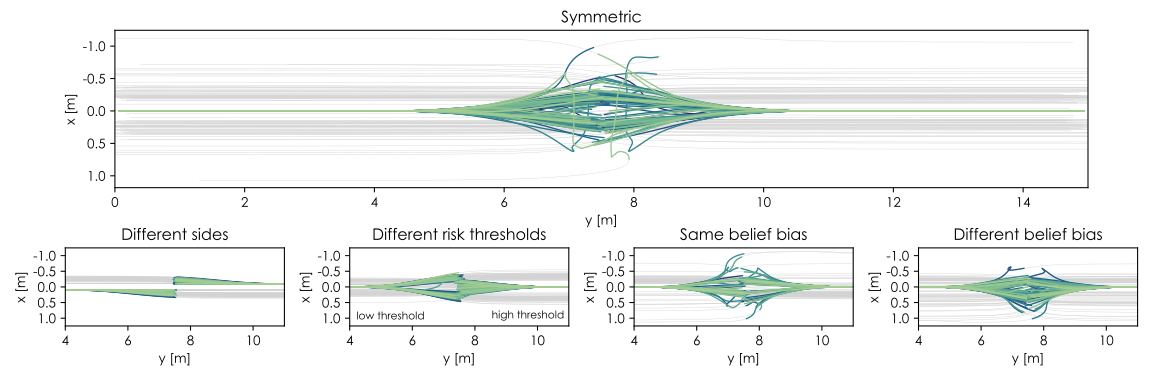}
    \caption{Position traces for all 500 model simulations per scenario, rotated $90\deg$ clockwise (in comparison to figures~\ref{fig:introduction-overview} and~\ref{fig:example_interaction}). The colored lines represent the pedestrians' positions up to the point where they pass each other. Thin gray lines represent the remainder of the traces. The colors indicate different trials. The colored triangles in the "Symmetric" plot indicate the starting positions of the orange and blue pedestrians.}
    \label{fig:traces}
\end{figure*}

\begin{figure*}[t]
    \centering
    \includegraphics[width=1.\linewidth]{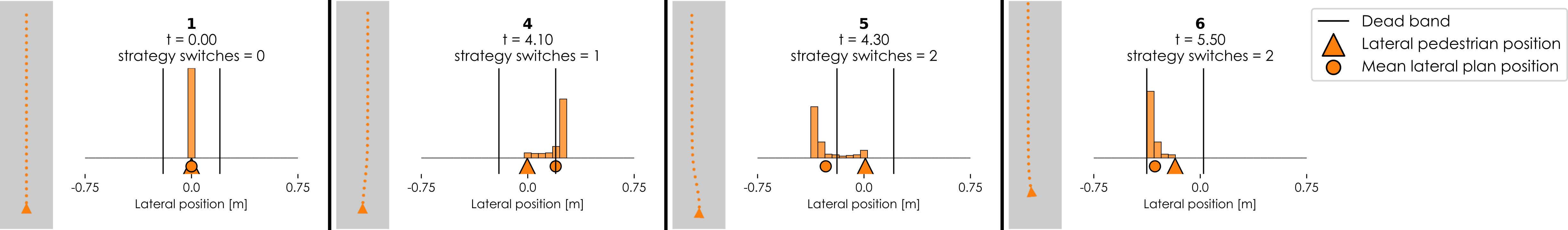}
    \caption{An illustration of how the number of strategy switches is determined, with the orange pedestrian's plan from Figure~\ref{fig:example_interaction}. A strategy switch is defined as a plan update where the plan changes from moving to one side (e.g., to the left) to moving to the other side (the right). Every pedestrian has a lateral dead band of $+-0.2~m$ (black lines) around their current position (triangles). The mean x-position (circles) of all plan points (histograms) determines this direction. The initial mean x-value lies within the dead band (frame 1). One strategy switch is counted once it moves outside the dead band (frame 4). When the average planned x-position moves to the other side of the dead band, another switch is counted (frame 5). However, if the average x value of the plan stays on that side or moves back inside the dead band, nothing happens (frame 6). Strategy switches are counted up until the moment when pedestrians pass each other. Therefore, pedestrians with zero strategy switches never substantially changed their plans; with one switch, they decided to go left or right and stuck with that, and with more switches, they adjusted their (high-level) plan multiple times.}
    \label{fig:straty_switches_def}
\end{figure*}

The position traces for all 500 simulations are shown in Figure~\ref{fig:traces}. For the symmetric scenario, both pedestrians move to both sides of the sidewalk to avoid collisions. The traces show simulations without a conflict and simulations where the pedestrians move sideways and struggle to pass each other. On the contrary, no conflicts are visible in the scenario where pedestrians start with a lateral offset. Sometimes, pedestrians move to the side of the sidewalk to provide extra space for the passing pedestrian and then continue straight. This behavioral pattern has previously been observed in real-world pedestrian encounters between few pedestrians (i.e., excluding crowds)~\cite{Kim2014}. The scenario with different risk thresholds also shows fewer conflicts than the symmetric scenario. In most cases, the pedestrian with the lower risk threshold (coming from the left) moves out of their way to avoid conflict. In contrast, the other pedestrian (coming from the right) does not always deviate from their initial path. In the simulations where they do deviate, they deviate later than the pedestrian with the low-risk threshold.

When the pedestrians have the same belief bias, they move mostly to the right to pass each other. However, some conflicts still occur. In the scenario with different belief biases, it can be clearly seen that both pedestrians move to the same side in many cases (in the negative x direction). This results in many conflicts. The number of strategy switches for each pedestrian per trial can provide more insight into the number of conflicts and sidewalk salsas in the different scenarios.

\begin{figure}
    \centering
    \includegraphics[width=0.8\linewidth]{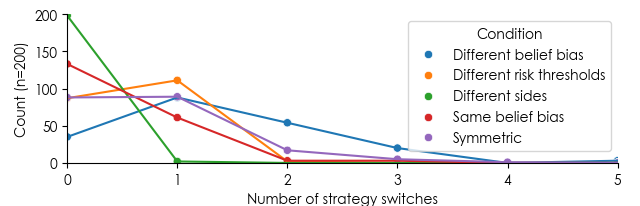}
    \caption{The number of strategy switches a single pedestrian made in a single trial per condition. $n=200$ since there are $100$ trials per condition with $2$ pedestrians in each trial.}
    \label{fig:straty_switches_per_condition}
\end{figure}

The number of strategy switches represents how often individual pedestrians had to change their high-level plan (i.e., going to the left or right) to come to a safe solution (Figure~\ref{fig:straty_switches_def}). This metric reveals that pedestrians in the "different sides" and "same belief bias" scenarios often did not change their plan at all; they continued their initial plan with zero strategy switches (Figure~\ref{fig:straty_switches_per_condition}). For the other scenarios, pedestrians mostly changed their plan once, indicating an easy solution. In the "different belief bias" scenario, pedestrians needed two or more strategy switches most often.

These strategy switches provide insight into the individual behaviors of the pedestrians, but the sidewalk salsa is a joint effort. To investigate the number of sidewalk salsas per condition, I defined the occurrence of the sidewalk salsa as the case where both pedestrians performed at least two strategy switches (Table~\ref{tab:end_states}). This data shows that the model can reproduce the sidewalk salsa under specific (symmetrical) conditions while minor positional or personal changes reduce the number of sidewalk salsas to 0. Furthermore, the model can represent biases in the beliefs of pedestrians. These simple biases (i.e., adjustments to the weight of the belief that someone is going left or right) reduce the number of sidewalk salsas if they are compatible and substantially increase the number if they are opposing.

Finally, all simulated data was used in this analysis, but not all simulations finished with one of the pedestrians reaching the other side of the sidewalk. In a limited number of cases, the pedestrian collided (Table~\ref{tab:end_states}). A large majority (25) of simulations that ended in a collision contained a sidewalk salsa. Intuitively, the number of collisions might seem high, especially in the "different belief bias" scenario. This could be explained by the fact that pedestrians in this simulation have no means of communication other than their movements. In real life, people will most likely resort to other means of communication --such as hand gestures or speech ("After you!")-- when they find themselves in a severe instance of the sidewalk salsa. These behaviors are not possible in the simulation, and thus, the only option is to keep trying to resolve the conflict and possibly collide.

\begin{table}[h]
    \centering
    \caption{End states and sidewalk salsas per condition}
    \label{tab:end_states}
    \begin{tabular}{p{2.8cm}|c c c} 
         & Finished & Collided &  Sidewalk Salsas \\ \hline
         Symmetric & 98 & 2 &  6 \\
         Different sides & 100 & 0 & 0 \\
         Different risk thresholds & 100 & 0 & 0 \\
         Same belief bias & 100 & 0 & 3  \\
         Different belief bias & 75 & 25  & 26 \\ 
    \end{tabular}
\end{table}

\section{Discussion}
In this work, I presented a model that can reproduce the sidewalk salsa under specific circumstances, based on the Communication-Enabled Interaction framework~\cite{Siebinga2023}. In 500 model simulations with two pedestrians approaching each other on a sidewalk, the sidewalk salsa occurred in symmetric situations where both pedestrians have the same risk thresholds and approach each other head-on. Slight deviations in personal preferences (risk thresholds) or initial lateral positions reduced the number of sidewalk salsas. In the symmetric scenario, biases in the belief about on which side the other pedestrian will pass lower the number of sidewalk salsas when both pedestrians share the same bias but increase this number when pedestrians have opposing biases. This provides evidence for the hypothesis that cultural norms --based on the side of the road on which cars drive-- influence the occurrence of the sidewalk salsa~\cite{Munro2019, Holohan2018, Pfaff2018, Moussaid2009}.

Although this is the first model explicitly targeted at replicating the sidewalk salsa, a lot of research has been done on pedestrian interactions, both in the real world (e.g.~\cite{Kim2014, Turnwald2014, Turnwald2016, Salzmann2020}) and in laboratory settings (e.g.~\cite{Gorrini2014, Olivier2012}). The simulated behavior of the model in interactions that did not end in the sidewalk salsa corresponds to phenomena observed in these previous studies. In 2014, Kim et al. analyzed videos of pedestrian interactions~\cite{Kim2014} and concluded that during low pedestrian flow (i.e., interactions between few pedestrians), typically only one pedestrian makes an evasive movement. This corresponds to the model's behavior in the scenarios with different risk thresholds and different sides. However, on some occasions, both pedestrians make evasive movements; this also happens occasionally in the simulations in these scenarios. Finally, during low pedestrian flow, pedestrians typically do not return to their previous path (i.e., their initial lateral position on the sidewalk); this corresponds to the model's simulations (Figure~\ref{fig:traces}).

Comparing the model with previous works also reveals some opportunities for improvement. Pedestrians in earlier works mostly seem to "steer" their trajectory during walking (e.g.,~\cite{Olivier2012}) and thus adhere to non-holonomic dynamics. Therefore, the model presented in this paper uses headings as a means of communication. However, since sidestepping (holonomic dynamics) can also be expected during a sidewalk salsa, the pedestrians adhere to a dynamics model that allows for both holonomic and non-holonomic movements. Whether pedestrians are best modeled with holonomic or non-holonomic dynamics (see~\cite{Turnwald2014, Turnwald2016} for examples and a comparison of both) is still an open question and might depend strongly on the purpose of a model. Although the model here describes plausible pedestrian motions in most cases, it assumes that forward and orthogonal movements are controlled through acceleration inputs. However, while side-stepping, pedestrians are likely to take only one or two steps in a lateral direction, longer lateral movements are unlikely. Therefore, this control assumption might lead to unrealistically long lateral movements (observed in a few trials in Figure~\ref{fig:traces}). Thus, more work is needed to investigate the dynamic behavior of pedestrians in high-risk scenarios, such as during the sidewalk salsa, and how that is best captured in a dynamic model.

The parameters for walking speed $v$, expected lateral acceleration $a_e$, and comfortable range $r_{com}$ were based on previous studies~\cite{Zebala2012, Gorrini2014, Kim2014}. However, the model simulations only considered single values for these parameters, while the literature suggests that there might be inter-dependencies between them. In 2014, Gorrini et al. concluded that walking speed significantly affects personal space (i.e., the comfortable range)~\cite{Gorrini2014}. Therefore, it might be necessary to implement dynamics values for $r_{com}$ when simulating the model for different walking speeds. In 2012, Z\c{e}bala et al. showed that walking acceleration and speed depend on observable features such as age, gender, and body mass index~\cite{Zebala2012}. Therefore, pedestrians might adjust their expectations about the accelerations and speeds of other pedestrians based on these features, something that could be investigated with the model as well.

Many previous works on pedestrian behavior collected trajectory data on pedestrian interactions either in the lab or real-world situations (e.g.~\cite{Kim2014, Lerner2007, Pellegrini2009, Olivier2012}). However, none of these studies report instances of the sidewalk salsa in their data. So, although the phenomenon is well known, it is rarely captured. One researcher even said in an interview that he tried to replicate the sidewalk salsa in a lab experiment but did not succeed~\cite{Holohan2018}. This lack of data and the difficulty of capturing the sidewalk salsa make validating the proposed model on (existing) real data difficult, if not impossible. Future work could investigate how the sidewalk salsa could be captured or reproduced in a lab environment or in the real world. 

Despite these limitations, the model could prove valuable in enhancing our understanding of the finer nuances in pedestrian interactions and in applications for mobile robotics. To start with the first, the model showed it can replicate the sidewalk salsa under specific circumstances. It can help to identify further the conditions under which these awkward interactions happen and provide insight into the effects of these conditions on the (modeled) internal mechanisms underlying the interactions. When tuned and validated on empirical data, it could help identify how external factors and movements of pedestrians (or autonomous agents) influence the beliefs and risk perception of pedestrians, aspects that are normally difficult to measure.

Research on mobile robot navigation and their interactions with pedestrians has so far mainly focussed on the freezing robot problem (e.g.~\cite{Trautman2010}), efficient crowd navigation (e.g.~\cite{Chen2018}), or negotiating around obstacles and collision-free planning (e.g.~\cite{Turnwald2014, Dadvar2021}). None of these examples consider individual interactions with pedestrians, while these frequently occur in real-world applications~\cite{Weinberg2023, Gehrke2023, JeredVroon2020}. 

Therefore, designing mobile robots that can safely and efficiently interact with individual humans is an important open research problem. Inspiration for how to approach this problem can be taken from autonomous vehicle research, where substantial effort has gone into investigating how to interact with individual humans. One approach, the so-called interaction-aware controller~\cite{Siebinga2022}, uses driver behavior models to inform the robots in planning and decision-making. However, incorporating non-game-theory-based models into robot controllers is still an open problem. Another approach is to use driver models to develop and test new robot behavior~\cite{Li2018, Queiroz2022}. Pedestrian models capable of describing the nuances in the interaction between individual pedestrians and the underlying mechanisms that lead to the sidewalk salsa could be helpful for mobile robots in the same manner; for example, by extending existing tools for bench-marking with reactive pedestrian behavior~\cite{Biswas2022}. The proposed model could thus be a useful asset in developing social navigation for robots.

\section{Conclusion}
In this paper, I've proposed a model for pedestrian interactions on a sidewalk based on the Communication-Enabled Interaction framework~\cite{Siebinga2023}. I conclude the following:

\begin{itemize}
    \item The proposed model based on the Communication-Enabled Interaction framework can successfully reproduce the "Sidewalk Salsa" under certain symmetrical conditions.
    \item The model indicates that positional and personal differences between interacting pedestrians reduce the probability of a "Sidewalk Salsa" occurring; furthermore, the model shows behavior comparable to previously observed pedestrian behavior in these scenarios without sidewalk salsas.
    \item The model shows that interactions between pedestrians with compatible belief biases are less likely to result in a "Sidewalk Salsa" than interactions where pedestrians have opposing belief biases, and thereby supports the hypothesis that cultural norms indicating on which side to pass play a role in the occurrence of the sidewalk salsa~\cite{Munro2019, Holohan2018, Pfaff2018, Moussaid2009}.
\end{itemize}

\section*{Acknowledgments}
I thank David Abbink and Arkady Zgonnikov for their valuable feedback on the figures and the manuscript.

\bibliographystyle{ieeetr}
\bibliography{my_collection.bib}

\begin{thebibliography}{10}

\bibitem{Andersson2018}
J.~A.~E. Andersson, J.~Gillis, G.~Horn, J.~B. Rawlings, and M.~Diehl, ``{CasADi} -- {A} software framework for nonlinear optimization and optimal control,'' {\em Mathematical Programming Computation}, 2018.

\bibitem{Siebinga2024b}
O.~Siebinga, ``{Data underlying the publication "A Model of the Sidewalk Salsa"},'' 2024.

\bibitem{Holohan2018}
M.~Holohan, ``{Researchers explain that awkward sidewalk dance we all do — and hate},'' 2018.

\bibitem{Bugler1997}
B.~P. Bugler, ``{Sitting Out the Sidewalk Dance},'' 1997.

\bibitem{Burgress2019}
G.~Burgress, ``{Science behind 'footpath foxtrot' or 'sidewalk shuffle' linked to brain's ability to predict future},'' 2019.

\bibitem{LaBelle2015b}
C.~LaBelle, ``{Pavement Polka},'' 2015.

\bibitem{Munro2019}
E.~Munro, ``{THE I'M-TRYING-TO-GET-AROUND-YOU DANCE EXPLAINED},'' 2019.

\bibitem{LaBelle2015a}
C.~LaBelle, ``{Sidewalk Salsa},'' 2015.

\bibitem{Gehrke2023}
S.~R. Gehrke, C.~D. Phair, B.~J. Russo, and E.~J. Smaglik, ``{Observed sidewalk autonomous delivery robot interactions with pedestrians and bicyclists},'' {\em Transportation Research Interdisciplinary Perspectives}, vol.~18, no.~March, p.~100789, 2023.

\bibitem{Weinberg2023}
D.~Weinberg, H.~Dwyer, S.~E. Fox, and N.~Martelaro, ``{Sharing the Sidewalk: Observing Delivery Robot Interactions with Pedestrians during a Pilot in Pittsburgh, PA},'' {\em Multimodal Technologies and Interaction}, vol.~7, no.~5, 2023.

\bibitem{Dadvar2021}
M.~Dadvar, K.~Majd, E.~Oikonomou, G.~Fainekos, and S.~Srivastava, ``{Joint Communication and Motion Planning for Cobots},'' in {\em Proceedings - IEEE International Conference on Robotics and Automation}, pp.~4771--4777, sep 2022.

\bibitem{Chen2018}
Z.~Chen, C.~Jiang, and Y.~Guo, ``{Pedestrian-Robot Interaction Experiments in an Exit Corridor},'' {\em 2018 15th International Conference on Ubiquitous Robots, UR 2018}, pp.~29--34, 2018.

\bibitem{Fang_Mei_Yuan_Wang_Wang_Wang_2021}
B.~Fang, G.~Mei, X.~Yuan, L.~Wang, Z.~Wang, and J.~Wang, ``Visual slam for robot navigation in healthcare facility,'' {\em Pattern Recognition}, vol.~113, p.~107822, May 2021.

\bibitem{Möller_Furnari_Battiato_Härmä_Farinella_2021}
R.~Möller, A.~Furnari, S.~Battiato, A.~Härmä, and G.~M. Farinella, ``A survey on human-aware robot navigation,'' {\em Robotics and Autonomous Systems}, vol.~145, p.~103837, Nov. 2021.

\bibitem{Guest2021}
O.~Guest and A.~E. Martin, ``{How Computational Modeling Can Force Theory Building in Psychological Science},'' {\em Perspectives on Psychological Science}, vol.~16, no.~4, pp.~789--802, 2021.

\bibitem{Siebinga2022}
O.~Siebinga, A.~Zgonnikov, and D.~Abbink, ``{A Human Factors Approach to Validating Driver Models for Interaction-aware Automated Vehicles},'' {\em ACM Transactions on Human-Robot Interaction}, vol.~11, pp.~1--21, dec 2022.

\bibitem{Honma2015}
M.~Honma, S.~Koyama, and M.~Kawamura, ``{Hesitant avoidance while walking: an error of social behavior generated by mutual interaction},'' {\em Frontiers in Psychology}, vol.~6, no.~July, pp.~1--8, 2015.

\bibitem{Nummenmaa2009}
L.~Nummenmaa, J.~Hy{\"{o}}n{\"{a}}, and J.~K. Hietanen, ``{I'll Walk This Way: Eyes Reveal the Direction of Locomotion and Make Passersby Look and Go the Other Way},'' {\em Psychological Science}, vol.~20, pp.~1454--1458, dec 2009.

\bibitem{Gorrini2014}
A.~Gorrini, K.~Shimura, S.~Bandini, K.~Ohtsuka, and K.~Nishinari, ``{Experimental Investigation of Pedestrian Personal Space},'' {\em Transportation Research Record: Journal of the Transportation Research Board}, vol.~2421, no.~1, pp.~57--63, 2014.

\bibitem{Olivier2012}
A.~H. Olivier, A.~Marin, A.~Cr{\'{e}}tual, and J.~Pettr{\'{e}}, ``{Minimal predicted distance: A common metric for collision avoidance during pairwise interactions between walkers},'' {\em Gait and Posture}, vol.~36, no.~3, pp.~399--404, 2012.

\bibitem{Gerin-Lajoie2005}
M.~G{\'{e}}rin-Lajoie, C.~L. Richards, and B.~J. McFadyen, ``{The negotiation of stationary and moving obstructions during walking: Anticipatory locomotor adaptations and preservation of personal space},'' {\em Motor Control}, vol.~9, no.~3, pp.~242--269, 2005.

\bibitem{Vassallo2018}
C.~Vassallo, A.~H. Olivier, P.~Sou{\`{e}}res, A.~Cr{\'{e}}tual, O.~Stasse, and J.~Pettr{\'{e}}, ``{How do walkers behave when crossing the way of a mobile robot that replicates human interaction rules?},'' {\em Gait and Posture}, vol.~60, no.~November 2017, pp.~188--193, 2018.

\bibitem{Pfaff2018}
L.~M. Pfaff and M.~E. Cinelli, ``{Avoidance behaviours of young adults during a head-on collision course with an approaching person},'' {\em Experimental Brain Research}, vol.~236, no.~12, pp.~3169--3179, 2018.

\bibitem{Moussaid2009}
M.~Moussa{\"{i}}d, D.~Helbing, S.~Garnier, A.~Johansson, M.~Combe, and G.~Theraulaz, ``{Experimental study of the behavioural mechanisms underlying self-organization in human crowds},'' {\em Proceedings of the Royal Society B: Biological Sciences}, vol.~276, no.~1668, pp.~2755--2762, 2009.

\bibitem{Vassallo2017}
C.~Vassallo, A.~H. Olivier, P.~Sou{\`{e}}res, A.~Cr{\'{e}}tual, O.~Stasse, and J.~Pettr{\'{e}}, ``{How do walkers avoid a mobile robot crossing their way?},'' {\em Gait and Posture}, vol.~51, pp.~97--103, 2017.

\bibitem{Lerner2007}
A.~Lerner, Y.~Chrysanthou, and D.~Lischinski, ``{Crowds by example},'' {\em Computer Graphics Forum}, vol.~26, no.~3, pp.~655--664, 2007.

\bibitem{Pellegrini2009}
S.~Pellegrini, A.~Ess, K.~Schindler, and L.~{Van Gool}, ``{You'll never walk alone: Modeling social behavior for multi-target tracking},'' {\em Proceedings of the IEEE International Conference on Computer Vision}, no.~Iccv, pp.~261--268, 2009.

\bibitem{Biswas2022}
A.~Biswas, A.~Wang, G.~Silvera, A.~Steinfeld, and H.~Admoni, ``{SocNavBench: A Grounded Simulation Testing Framework for Evaluating Social Navigation},'' {\em ACM Transactions on Human-Robot Interaction}, vol.~11, no.~3, 2022.

\bibitem{Hoogendoorn_Daamen_2005}
S.~P. Hoogendoorn and W.~Daamen, ``Pedestrian behavior at bottlenecks,'' {\em Transportation Science}, vol.~39, p.~147–159, May 2005.

\bibitem{Echeverria-Huarte2023}
I.~Echeverr{\'{i}}a-Huarte and A.~Nicolas, ``{Body and mind: Decoding the dynamics of pedestrians and the effect of smartphone distraction by coupling mechanical and decisional processes},'' {\em Transportation Research Part C: Emerging Technologies}, vol.~157, no.~September, p.~104365, 2023.

\bibitem{Brito2020}
B.~Brito, H.~Zhu, W.~Pan, and J.~Alonso-Mora, ``{Social-VRNN: One-Shot Multi-modal Trajectory Prediction for Interacting Pedestrians},'' in {\em Proceedings of Machine Learning Research}, vol.~155, pp.~862--872, 2020.

\bibitem{Siebinga2023}
O.~Siebinga, A.~Zgonnikov, and D.~A. Abbink, ``{Modelling communication-enabled traffic interactions},'' {\em Royal Society Open Science}, vol.~10, may 2023.

\bibitem{Mombaur_Truong_Laumond_2010}
K.~Mombaur, A.~Truong, and J.-P. Laumond, ``From human to humanoid locomotion—an inverse optimal control approach,'' {\em Autonomous Robots}, vol.~28, p.~369–383, Apr. 2010.

\bibitem{Arechavaleta2008}
G.~Arechavaleta, J.~P. Laumond, H.~Hicheur, and A.~Berthoz, ``{On the nonholonomic nature of human locomotion},'' {\em Autonomous Robots}, vol.~25, no.~1-2, pp.~25--35, 2008.

\bibitem{Zebala2012}
J.~Zebala, P.~Ciepka, and A.~Reza, ``{Pedestrian acceleration and speeds},'' {\em Problems of Forensic Sciences}, vol.~91, pp.~227--234, 2012.

\bibitem{Siebinga2024}
O.~Siebinga, A.~Zgonnikov, and D.~Abbink, ``{A merging interaction model explains human drivers' behaviour from input signals to decisions},'' pp.~1--18, dec 2023.

\bibitem{Kim2014}
S.~Kim, J.~Choi, S.~Kim, and R.~Tay, ``{Personal space, evasive movement and pedestrian level of service},'' {\em Journal of Advanced Transportation}, vol.~48, pp.~673--684, oct 2014.

\bibitem{Turnwald2014}
A.~Turnwald, W.~Olszowy, D.~Wollherr, and M.~Buss, ``{Interactive navigation of humans from a game theoretic perspective},'' {\em IEEE International Conference on Intelligent Robots and Systems}, no.~Iros, pp.~703--708, 2014.

\bibitem{Turnwald2016}
A.~Turnwald, D.~Althoff, D.~Wollherr, and M.~Buss, ``{Understanding Human Avoidance Behavior: Interaction-Aware Decision Making Based on Game Theory},'' {\em International Journal of Social Robotics}, vol.~8, no.~2, pp.~331--351, 2016.

\bibitem{Salzmann2020}
T.~Salzmann, B.~Ivanovic, P.~Chakravarty, and M.~Pavone, ``{Trajectron++: Dynamically-Feasible Trajectory Forecasting with Heterogeneous Data},'' in {\em Lecture Notes in Computer Science (including subseries Lecture Notes in Artificial Intelligence and Lecture Notes in Bioinformatics)}, vol.~12363 LNCS, pp.~683--700, 2020.

\bibitem{Trautman2010}
P.~Trautman and A.~Krause, ``{Unfreezing the robot: Navigation in dense, interacting crowds},'' {\em IEEE/RSJ 2010 International Conference on Intelligent Robots and Systems, IROS 2010 - Conference Proceedings}, pp.~797--803, 2010.

\bibitem{JeredVroon2020}
{Jered Vroon}, {Zolt{\'{a}}n Rus{\'{a}}k}, and {Gerd Kortuem}, ``{Context-Confrontation: Elicitation and Exploration of Conflicts for Delivery Robots on Sidewalks},'' {\em First international workshop on Designerly HRI Knowledge: Held in conjunction with the 29th IEEE International Conference on Robot and Human Interactive Communication (RO-MAN 2020)}, vol.~2020, 2020.

\bibitem{Li2018}
N.~Li, D.~W. Oyler, M.~Zhang, Y.~Yildiz, I.~Kolmanovsky, and A.~R. Girard, ``{Game theoretic modeling of driver and vehicle interactions for verification and validation of autonomous vehicle control systems},'' {\em IEEE Transactions on Control Systems Technology}, vol.~26, no.~5, pp.~1782--1797, 2018.

\bibitem{Queiroz2022}
R.~Queiroz, D.~Sharma, R.~Caldas, K.~Czarnecki, S.~Garc{\'{i}}a, T.~Berger, and P.~Pelliccione, ``{A Driver-Vehicle Model for ADS Scenario-based Testing},'' pp.~1--16, may 2022.

\end{thebibliography}

\end{document}